%% file: simplification.tex
\documentclass[11pt]{article}
\usepackage{eacl2017}
\usepackage{times}
\usepackage{url}
\usepackage{latexsym}
\usepackage{amsmath}
\usepackage{graphicx}
\usepackage{tikz}

\usepackage[utf8]{inputenc}
\usepackage[T1]{fontenc}

\eaclfinalcopy % Uncomment this line for the final submission
\title{Neural Machine Translation from Simplified Translations}

\author{
  Josep Crego \ \  {\normalfont and} \ \  Jean Senellart \\
  {\tt firstname.lastname@systrangroup.com} \\
  SYSTRAN / 5 rue Feydeau, 75002 Paris, France
}

\date{}

\begin{document}
\maketitle
\begin{abstract}

Text simplification aims at reducing the lexical, grammatical and structural complexity of a text while keeping the same meaning. In the context of machine translation, we introduce the idea of simplified translations in order to boost the learning ability of deep neural translation models.
We conduct preliminary experiments showing that translation complexity is actually reduced in a translation of a source bi-text compared to the target reference of the bi-text while using a neural machine translation (NMT) system learned on the exact same bi-text.
Based on knowledge distillation idea, we then train an NMT system using the simplified bi-text, and show that it outperforms the initial system that was built over the reference data set. Performance is further boosted when both reference and automatic translations are used to learn the network. We perform an elementary analysis of the translated corpus and report accuracy results of the proposed approach on English-to-French and English-to-German translation tasks.

\end{abstract}

\section{Introduction}
\label{sec:intro}

Neural machine translation (NMT) has recently achieved state-of-the-art results in several translation tasks ~\cite{Bojar2016} and for various language pairs. Its conceptual simplicity has attracted many researchers as well as a growing number of private entities that have begun to include NMT engines in their production systems.

%Equivalently to its phrase-based predecessor,
NMT networks are directly learned from parallel bi-texts, consisting of large amounts of human sentences together with their corresponding translations.
Even if all translations in a bi-text are considered suitable, i.e. the meaning is preserved and the target language is fully correct, there is a large variability in these translations: in some cases the translations follow a more or less word-for-word pattern (literal translations), while in many others the translations are showing greater latitude of expression (free translations). A good human translation is often judged by this latitude of expressions. %Such "free" translations can also be seen as paraphrases of more literal translations.
In contrast, machine translations are usually "closer", in terms of syntactic structure and even word choices, to the input sentences. Hence, even when the translation output is very good, these translations are still generally closer to literal translations because "free translations" are by definition more complicated and less easy to learn and model. % It results difficult to find "free" translations from an automatic system.
It is also a rather intuitive idea that feeding more literal translation to a neural translation engine training should facilitate the training process compared to same training with less literal translations. 

We report preliminary results of experiments where we automatically simplify a human translation bi-text which is then used to train neural translation engines. %The rationale behind this experiments is to simplify the translation task,
Thus, boosting the learning ability of neural translation models and we show that the resulting models are performing even better than a neural translation engine trained on the reference dataset.
%Neural translation engines implement a beam search decoder where many different hypotheses are considered and the most likely, according to a deep neural network, is finally output.
The remaining of this paper is structured as follows.
Section \ref{sec:related} briefly surveys previous work.
Section \ref{sec:neural} outlines our neural MT engine.
Section \ref{sec:langsimp} details the simplification approach presented in this paper.
Section \ref{sec:exp} reports experimental results and section \ref{sec:conclusions} draws conclusion and proposes further work.

\section{Related Work}
\label{sec:related}

A neural encoder-decoder model performing text simplification at lexical and syntactic levels is reported in \cite{DBLP:journals/corr/WangCAQ16}. The work introduces a model for text simplification. However, it differs from our work in that we use a neural MT engine to simplify translations, which are further used to boost translation performance, while their end goal is text simplification. 
In \cite{degispert-iglesias-byrne:2015:NAACL-HLT}, a phrase-based SMT system is presented that employs as preprocessing module a neural network that models source-side preordering, aiming at finding a permutation of the source sentence that matches the target sentence word order. Also, with the objective of simplifying the translation task.
The work by \cite{DBLP:journals/corr/NiehuesCHW16} presents a technique to combine phrase-based and neural MT. The phrase-based system is initially used to produce a first hypothesis which is then considered together with the input sentence by a neural MT engine to produce the final hypothesis. The authors claim that the combined approach shows the strength of both approaches, namely fluent translations and the ability to translate rare words.

In this work we have used one of the knowledge distillation techniques detailed by \cite{knowledgedistillation2016} where the authors train a smaller student network to perform better by learning from a larger teacher network allowing to build more compact neural MT models. With a similar objective, \cite{2015arXiv150302531H} claim that distillation works well for transferring knowledge from an ensemble or from a large highly regularised model into a smaller, distilled model.

\section{Neural MT}
\label{sec:neural}

Our NMT system follows the architecture presented in ~\cite{DBLP:journals/corr/BahdanauCB14}. It is implemented as an encoder-decoder network with multiple layers of a RNN with Long Short-Term Memory hidden units ~\cite{DBLP:journals/corr/ZarembaSV14}.

The encoder is a bidirectional neural network that reads an input sequence $s = (s_1,...,s_J)$ and calculates a forward sequence of hidden states $(\overrightarrow{h_1}, ..., \overrightarrow{h_J})$, and a backward sequence $(\overleftarrow{h_1},..., \overleftarrow{h_J})$. The decoder is a RNN that predicts a target sequence $t = (t_1, ..., t_I)$, being $J$ and $I$ respectively the source and target sentence lengths. Each word $t_i$ is predicted based on a recurrent hidden state $h_i$, the previously predicted word $t_{i-1}$, and a context vector $c_i$. We employ the attentional architecture from ~\cite{luong-pham-manning:2015:EMNLP}.
The framework is available on the open-source project \texttt{seq2seq-attn}\footnote{\url{http://nlp.seas.harvard.edu}}. Additional details are given in \cite{DBLP:journals/corr/CregoKKRYSABCDE16}.

%Training is performed on a parallel corpus.
%For translation, a beam search with small beam size is used.
%The project is actively maintained by the Harvard NLP group.

\section{Translation Simplification}
\label{sec:langsimp}

Translation simplification is based on the idea that any sentence may have multiple translations, all being equally suitable. Following this idea, and despite the fact that deep neural networks have achieved excellent performance on many difficult tasks, we are interested in keeping the translation task as simple as possible. Hence, for a training bi-text we are interested in translations having a similar structure as source sentences. The following example shows an English sentence translated into two distinct French sentences:

\begin{picture}(100,75)
\centering
\put(0,55){\it This deficiency discourages the practice.}
\put(90,40){$\Downarrow$}
\put(0,25){\it Ce d\'efaut a un effet dissuasif sur la pratique.}
\put(0,10){\it Cette insuffisance d\'ecourage la pratique.}
\end{picture}

\noindent Both French translations are suitable. However, the last French translation is closer in terms of sentence structure to its English counterpart. %When "close" translations are considered the translation task is simplified, and therefore boosted the learning process.

Producing "close" translations is the natural behaviour of Machine Translation systems. Hence, we use a neural MT system to simplify a translation bi-text. Similar to knowledge distillation, target language simplification is performed in 3 steps: 
\begin{enumerate}
  \setlength\itemsep{-0.3em}
\item train a teacher model with reference translations,
\item run beam search over the training set with the teacher model,
\item train the student network on this new dataset.
\end{enumerate}

In the next Section, we analyse the training data translated by beam search following step (2) using the models built in step (1).

\subsection{Translated Language Analysis}
\label{ssec:analysis}

Based on the NMT system outlined in Section \ref{sec:neural} and following the language simplification method previously outlined, we train English-to-French and English-to-German teacher networks as detailed in Section \ref{ssec:training}. 
Using such teacher models we translate the English-side of both training sets producing respectively German and French translation hypotheses. Aiming for a better understanding of the translated languages we conduct an elementary human analysis of the French and German hypotheses. We mainly observe that in many cases, translation hypotheses produced by the teacher systems consist of paraphrases of the reference translations. 
Such hypotheses are closer in terms of syntactic structure to the source sentences than reference translations. Examples in Table \ref{tab:examples} illustrate this fact. While both, {\it Ref} and {\it Hyp} translations can be considered equally good, {\it Hyp} translations are syntactically closer to the source sentence. 
In the first example, the reference translation replaces the verb {\it receiving} with the action of {\it communicating}, hence subject and indirect objects are switched. 
In the second example several rephrasings are observed: {\it [Si cette ratification n'a pas lieu $\sim$ En l'absence d'une telle ratification]} and finally {\it [la commission devrait \^etre invit\'ee $\sim$ il y aurait lieu d'inviter la commission]}. In both examples meaning is fully preserved and both sentences are naturally good.

\begin{table*}[h]
\begin{center}
\scalebox{0.90}{
\begin{tabular}{|rl|}
 \hline
Src: & The Secretary-General has received views from Denmark and Kazakhstan. \\
Ref: & $[$Le Danemark et le Kazakhstan$]_{subj}$ {\bf ont communiqu\'e} leurs vues au $[$Secr\'etaire g\'en\'eral$]_{ind}$. \\
Hyp & $[$Le Secr\'etaire g\'en\'eral$]_{ind}$ {\bf a re\c cu} les vues du $[$Danemark et du Kazakhstan$]_{subj}$. \\
\hline
 Src:  & If this ratification does not take place , the Commission should be called. \\
 Ref:  & {\bf En l'absence d'une telle} ratification , {\bf il y aurait lieu d'}inviter la Commission. \\
 Hyp: & {\bf Si cette} ratification {\bf n'a pas lieu} , la Commission {\bf devrait \^etre} invit\'ee. \\
\hline
\end{tabular}
}
\end{center}
\caption{\label{tab:examples} Examples of English-to-French translation simplification. }
\end{table*}

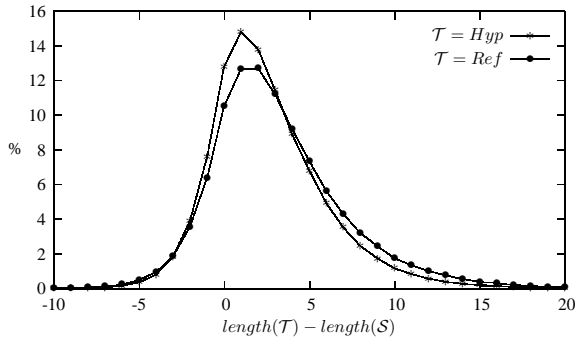
\begin{figure}[h]
\resizebox{\linewidth}{!}{\input{diffsize.tex}}
\caption{\label{fig:diffsize} English-to-French train set histogram representing difference in length between source $\mathcal{S}$ and target $\mathcal{T}$ sentences.}
\end{figure}

We conduct several experiments in order to confirm that translated hypotheses are closer to the input sentence than reference translations. We first measure the difference in length of {\it Hyp} and {\it Ref} translations with respect to the original source sentences $\mathcal{S}$. Figure \ref{fig:diffsize} shows the histogram for the English-to-French train set.
The number of target sentences $\mathcal{T}$ with similar length than source sentences $\mathcal{S}$ is higher for translated hypotheses $\mathcal{T}=Hyp$ than for reference translations $\mathcal{T}=Ref$. %Very similar results are obtained for the English-to-German data. %Note that the mean is not centered around $0$ since French sentences tend to be larger than their English translations.

\begin{figure}[h]
\resizebox{\linewidth}{!}{\input{crossings.tex}}
\caption{\label{fig:crossings} Difference in number of crossed alignments between machine and reference translations.}
\end{figure}
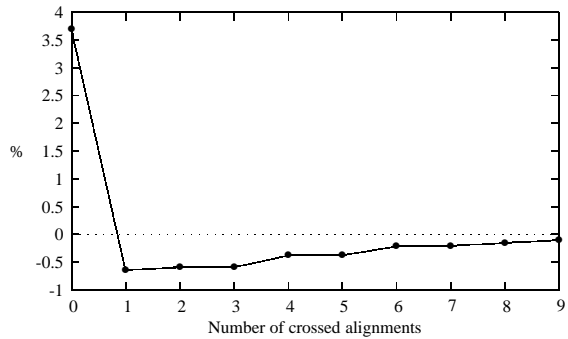

Additionally, we compare the number of crossed alignments\footnote{word alignments computed using \url{https://github.com/clab/fast_align}} on both language pairs (source-to-{\it Hyp} and source-to-{\it Ref}) in order to validate the closeness (similarity) of syntactic structures.
Given a sentence pair with its set of alignments, we compute for each source word $s_i$ the number of alignment crossings between the given source word and the rest of the source words. We consider that two alignments $(i,j)$ and $(i',j')$ are crossed if $(i-i')*(j-j') < 0$.
Figure \ref{fig:crossings} plots the difference in number of crossed alignments between source-to-{\it Hyp} and source-to-{\it Ref}.  As it can be seen, the source-to-{\it hyp} pair has a higher number of non-crossed alignments (near 4\%), while the number of words with crossed alignments is higher for the source-to-{\it Ref} pair. Statistics were computed over the same number of source words for both train pairs. %Similar results are obtained for the English-to-German data.
Notice that translated hypotheses {\it hyp} are automatically generated. Hence, carrying an important number of translation errors that cannot be neglected. The next Section evaluates the suitability of source-to-{\it hyp} translations as a train set for our neural MT systems compared to source-to-{\it ref} translations.

\section{Experiments}
\label{sec:exp}

We evaluate the presented approach on English-to-French and English-to-German translation.
%Section \ref{ssec:training} gives statistics of the corpora used and details training procedures, while Section \ref{ssec:results} reports on translation accuracy results.

\subsection{Training Details}
\label{ssec:training}

Experiments are performed using data made available for the shared translation task of the WMT\footnote{\url{http://www.statmt.org/wmt16/}}. %For English-to-French, we used {\it commoncrawl}, {\it europarl-v7}, {\it giga.release2}, {\it hansard}, {\it jrc\_2011}, {\it news-commentary-v8}, {\it OPUS-2013-Tatoeba}, {\it ted-talks-fbk} and {\it undoc.2000}. For English-to-German we used {\it europarl-v7}, {\it jrc\_2011}, {\it MultiUN}, {\it news-commentary-v11} and {\it ted-talks-fbk}.
Corpora is initially filtered using standard techniques to discard noisy translation examples. Tokenisation is also performed with an in-house toolkit, using standard token separators (spaces, tabs, etc.) as well as a set of language-dependent linguistic rules. %Several kinds of entities are recognised (url and number) replacing its content by the appropriate place-holder. %A post-process is used to detokenise translation hypotheses, where the original raw text format is regenerated following equivalent techniques. 
Corpora was split into three separate sets, Train, Validation and Test data sets. Finally, a random subset of the entire training set is kept containing $1$ million sentence pairs. Table \ref{tab:corpora} contains statistics for the data used in both translation tasks.
All experiments employ the NMT system detailed in Section \ref{sec:neural} on our NVidia GeForce GTX 1080. We keep the most frequent $50,000$ words for source and target vocabularies with a word embedding size of $500$ cells. During training we use stochastic gradient descent, a mini-batch size of $64$ with dropout probability set to $0.3$ and bidirectional RNN. We train our models for $18$ epochs. Learning rate is set to $1$ and start decay after epoch $10$ by $0.5$. For decoding, we always use a beam size of $5$.

\begin{table}[h]
\begin{center}
\scalebox{0.90}{
\begin{tabular}{|l|cc|cc|}
\hline
 & \bf En & \bf De & \bf En & \bf Fr \\
\hline
\multicolumn{5}{l}{Train} \\
\hline
Lines  & \multicolumn{2}{c|}{1M} & \multicolumn{2}{c|}{1M} \\
Words        & 25.7M & 25.2M & 24.4M & 27.6M \\
Vocab. & 151k & 249k     & 196k & 224k \\
\hline
\multicolumn{5}{l}{Validation} \\
\hline
Lines  & \multicolumn{2}{c|}{2,000} & \multicolumn{2}{c|}{2,000} \\
Words         & 51.2k & 50.4k & 48.9k & 55.6k \\
Vocab. & 7,183 & 9,837 & 8,168 & 8,773 \\
OOV           & 120    & 208    & 169    & 202 \\
\hline
\multicolumn{5}{l}{Test-intern} \\
\hline
Lines  & \multicolumn{2}{c|}{2,000} &  \multicolumn{2}{c|}{2,000} \\
Words         & 52.3k & 51.3k & 48.2k & 54.7k \\
Vocab. & 7,232 & 8,868 & 8,199 & 8,623 \\
OOV           & 107    & 195    & 203   &  222 \\
\hline
\multicolumn{5}{l}{newstest2014} \\
\hline
Lines  & \multicolumn{2}{c|}{3,003} &  \multicolumn{2}{c|}{3,003} \\
Words         & 69.7k & 86.9k & 72.2k & 70.1k \\
Vocab.        & 9,912 & 11,183 & 9,978 & 12,683 \\
OOV           & 1,597 & 956 & 856    & 1,761 \\
\hline
\end{tabular}
}
\end{center}
\caption{\label{tab:corpora} Statistics of Train, Validation and Test data sets of English-to-German and English-to-French. M and k stand for millions and thousands.}
\end{table}

\vspace{-15px}
\subsection{Results}
\label{ssec:results}

Table \ref{tab:results} summarises translation accuracy results.% for different network configurations.
The first two rows of each translation task show teacher networks built using respectively $2$ and $4$ layers. Using $4$ layers obtains slightly better results for the English-to-German task while no difference is observed for the French-to-English task. On the third row we see accuracy results of student networks, built by distillation of the networks in the second row. Note that student networks were built using half the number of layers than their corresponding teacher networks. Finally the last row show accuracies of networks built using both, reference (R) and automatic (A) translations. Note that this last configuration employs double the number of training sentences than previous ones. Student networks outperform their corresponding teacher networks in all cases. Difference in BLEU \cite{Papineni:2002:BMA:1073083.1073135} scores between student and teacher configurations are shown in parentheses. Performance is further boosted when using both reference and hypothesis translations to train the networks.

\begin{table}[h]
\begin{center}
\scalebox{0.9}{
\begin{tabular}{|cc|ll|}
\hline
\bf Data & \bf Layers & \bf Test-intern & \bf newstest2014 \\
\hline
\multicolumn{4}{l}{English-to-German} \\
\hline
R & 2 x 800      & 33.05  & 19.25  \\
\hline
R & 4 x 800     & 33.40 & 19.40 \\
A & 2 x 800     & 33.91 (+0.51) & 20.26 (+0.86) \\
%\hline
R+A & 4 x 800 & 34.59 (+1.19) & 21.84 (+2.44) \\  
\hline
\multicolumn{4}{l}{English-to-French} \\
\hline
R & 2 x 1000     &  53.67 & 32.15 \\
\hline
R & 4 x 1000     & 53.78 &  32.13 \\
A & 2 x 1000     & 54.47 (+0.69) & 32.85 (+0.72) \\
%\hline
R+A & 4 x 1000 &  55.24 (+1.46) & 33.47 (+1.34) \\  
\hline
\end{tabular}
}
\end{center}
\caption{\label{tab:results} BLEU scores over Test-intern and newstest2014 sets of both translation tasks according to different network configurations. BLEU scores were calculated using \url{multi-bleu.perl}.  }
\end{table}

\vspace{-15px}
\section{Conclusions}
\label{sec:conclusions}

We have presented translation simplification experiments for neural machine translation. Results indicate the suitability of using simplified translations to train neural MT systems. Higher accuracy results are obtained by the systems trained using simplified data. Further experiments need to be carried out to validate the presented approach on additional language pairs and under different data size conditions.

\section*{Acknowledgments}

We would like to thank Yoon Kim and Prof. Alexander Rush for their valuable insights with knowledge distillation experiments.

\bibliography{eacl2017}
\bibliographystyle{eacl2017}

\end{document}

%% file: diffsize.tex
% GNUPLOT: LaTeX picture
\setlength{\unitlength}{0.240900pt}
\ifx\plotpoint\undefined\newsavebox{\plotpoint}\fi
\sbox{\plotpoint}{\rule[-0.200pt]{0.400pt}{0.400pt}}%
\begin{picture}(1500,900)(0,0)
\sbox{\plotpoint}{\rule[-0.200pt]{0.400pt}{0.400pt}}%
\put(110.0,131.0){\rule[-0.200pt]{4.818pt}{0.400pt}}
\put(90,131){\makebox(0,0)[r]{ 0}}
\put(1419.0,131.0){\rule[-0.200pt]{4.818pt}{0.400pt}}
\put(110.0,222.0){\rule[-0.200pt]{4.818pt}{0.400pt}}
\put(90,222){\makebox(0,0)[r]{ 2}}
\put(1419.0,222.0){\rule[-0.200pt]{4.818pt}{0.400pt}}
\put(110.0,313.0){\rule[-0.200pt]{4.818pt}{0.400pt}}
\put(90,313){\makebox(0,0)[r]{ 4}}
\put(1419.0,313.0){\rule[-0.200pt]{4.818pt}{0.400pt}}
\put(110.0,404.0){\rule[-0.200pt]{4.818pt}{0.400pt}}
\put(90,404){\makebox(0,0)[r]{ 6}}
\put(1419.0,404.0){\rule[-0.200pt]{4.818pt}{0.400pt}}
\put(110.0,495.0){\rule[-0.200pt]{4.818pt}{0.400pt}}
\put(90,495){\makebox(0,0)[r]{ 8}}
\put(1419.0,495.0){\rule[-0.200pt]{4.818pt}{0.400pt}}
\put(110.0,586.0){\rule[-0.200pt]{4.818pt}{0.400pt}}
\put(90,586){\makebox(0,0)[r]{ 10}}
\put(1419.0,586.0){\rule[-0.200pt]{4.818pt}{0.400pt}}
\put(110.0,677.0){\rule[-0.200pt]{4.818pt}{0.400pt}}
\put(90,677){\makebox(0,0)[r]{ 12}}
\put(1419.0,677.0){\rule[-0.200pt]{4.818pt}{0.400pt}}
\put(110.0,768.0){\rule[-0.200pt]{4.818pt}{0.400pt}}
\put(90,768){\makebox(0,0)[r]{ 14}}
\put(1419.0,768.0){\rule[-0.200pt]{4.818pt}{0.400pt}}
\put(110.0,859.0){\rule[-0.200pt]{4.818pt}{0.400pt}}
\put(90,859){\makebox(0,0)[r]{ 16}}
\put(1419.0,859.0){\rule[-0.200pt]{4.818pt}{0.400pt}}
\put(110.0,131.0){\rule[-0.200pt]{0.400pt}{4.818pt}}
\put(110,90){\makebox(0,0){-10}}
\put(110.0,839.0){\rule[-0.200pt]{0.400pt}{4.818pt}}
\put(332.0,131.0){\rule[-0.200pt]{0.400pt}{4.818pt}}
\put(332,90){\makebox(0,0){-5}}
\put(332.0,839.0){\rule[-0.200pt]{0.400pt}{4.818pt}}
\put(553.0,131.0){\rule[-0.200pt]{0.400pt}{4.818pt}}
\put(553,90){\makebox(0,0){ 0}}
\put(553.0,839.0){\rule[-0.200pt]{0.400pt}{4.818pt}}
\put(775.0,131.0){\rule[-0.200pt]{0.400pt}{4.818pt}}
\put(775,90){\makebox(0,0){ 5}}
\put(775.0,839.0){\rule[-0.200pt]{0.400pt}{4.818pt}}
\put(996.0,131.0){\rule[-0.200pt]{0.400pt}{4.818pt}}
\put(996,90){\makebox(0,0){ 10}}
\put(996.0,839.0){\rule[-0.200pt]{0.400pt}{4.818pt}}
\put(1218.0,131.0){\rule[-0.200pt]{0.400pt}{4.818pt}}
\put(1218,90){\makebox(0,0){ 15}}
\put(1218.0,839.0){\rule[-0.200pt]{0.400pt}{4.818pt}}
\put(1439.0,131.0){\rule[-0.200pt]{0.400pt}{4.818pt}}
\put(1439,90){\makebox(0,0){ 20}}
\put(1439.0,839.0){\rule[-0.200pt]{0.400pt}{4.818pt}}
\put(110.0,131.0){\rule[-0.200pt]{0.400pt}{175.375pt}}
\put(110.0,131.0){\rule[-0.200pt]{320.156pt}{0.400pt}}
\put(1439.0,131.0){\rule[-0.200pt]{0.400pt}{175.375pt}}
\put(110.0,859.0){\rule[-0.200pt]{320.156pt}{0.400pt}}
\put(10,500){\makebox(0,0){\%}} 
\put(774,29){\makebox(0,0){$length(\mathcal{T}) - length(\mathcal{S})$}} %%%%%%%%%%%%%%%%
\put(1279,733){\makebox(0,0)[r]{$\mathcal{T}=Ref$}} %%%%%%%%%%%%%%%%
\put(1299.0,733.0){\rule[-0.200pt]{24.090pt}{0.400pt}} %%%%%%%%%%%%%%%%%
\put(110,132){\usebox{\plotpoint}}
\put(110,131.67){\rule{10.600pt}{0.400pt}}
\multiput(110.00,131.17)(22.000,1.000){2}{\rule{5.300pt}{0.400pt}}
\put(154,132.67){\rule{10.840pt}{0.400pt}}
\multiput(154.00,132.17)(22.500,1.000){2}{\rule{5.420pt}{0.400pt}}
\multiput(199.00,134.61)(9.616,0.447){3}{\rule{5.967pt}{0.108pt}}
\multiput(199.00,133.17)(31.616,3.000){2}{\rule{2.983pt}{0.400pt}}
\multiput(243.00,137.59)(4.829,0.477){7}{\rule{3.620pt}{0.115pt}}
\multiput(243.00,136.17)(36.487,5.000){2}{\rule{1.810pt}{0.400pt}}
\multiput(287.00,142.58)(1.918,0.492){21}{\rule{1.600pt}{0.119pt}}
\multiput(287.00,141.17)(41.679,12.000){2}{\rule{0.800pt}{0.400pt}}
\multiput(332.00,154.58)(1.055,0.496){39}{\rule{0.938pt}{0.119pt}}
\multiput(332.00,153.17)(42.053,21.000){2}{\rule{0.469pt}{0.400pt}}
\multiput(376.00,175.58)(0.549,0.498){77}{\rule{0.540pt}{0.120pt}}
\multiput(376.00,174.17)(42.879,40.000){2}{\rule{0.270pt}{0.400pt}}
\multiput(420.58,215.00)(0.498,0.889){85}{\rule{0.120pt}{0.809pt}}
\multiput(419.17,215.00)(44.000,76.321){2}{\rule{0.400pt}{0.405pt}}
\multiput(464.58,293.00)(0.498,1.440){87}{\rule{0.120pt}{1.247pt}}
\multiput(463.17,293.00)(45.000,126.412){2}{\rule{0.400pt}{0.623pt}}
\multiput(509.58,422.00)(0.498,2.137){85}{\rule{0.120pt}{1.800pt}}
\multiput(508.17,422.00)(44.000,183.264){2}{\rule{0.400pt}{0.900pt}}
\multiput(553.58,609.00)(0.498,1.118){85}{\rule{0.120pt}{0.991pt}}
\multiput(552.17,609.00)(44.000,95.943){2}{\rule{0.400pt}{0.495pt}}
\put(597,706.67){\rule{10.840pt}{0.400pt}}
\multiput(597.00,706.17)(22.500,1.000){2}{\rule{5.420pt}{0.400pt}}
\multiput(642.58,705.02)(0.498,-0.774){85}{\rule{0.120pt}{0.718pt}}
\multiput(641.17,706.51)(44.000,-66.509){2}{\rule{0.400pt}{0.359pt}}
\multiput(686.58,636.15)(0.498,-1.038){85}{\rule{0.120pt}{0.927pt}}
\multiput(685.17,638.08)(44.000,-89.075){2}{\rule{0.400pt}{0.464pt}}
\multiput(730.58,545.49)(0.498,-0.936){87}{\rule{0.120pt}{0.847pt}}
\multiput(729.17,547.24)(45.000,-82.243){2}{\rule{0.400pt}{0.423pt}}
\multiput(775.58,461.64)(0.498,-0.889){85}{\rule{0.120pt}{0.809pt}}
\multiput(774.17,463.32)(44.000,-76.321){2}{\rule{0.400pt}{0.405pt}}
\multiput(819.58,384.32)(0.498,-0.682){85}{\rule{0.120pt}{0.645pt}}
\multiput(818.17,385.66)(44.000,-58.660){2}{\rule{0.400pt}{0.323pt}}
\multiput(863.58,324.70)(0.498,-0.568){85}{\rule{0.120pt}{0.555pt}}
\multiput(862.17,325.85)(44.000,-48.849){2}{\rule{0.400pt}{0.277pt}}
\multiput(907.00,275.92)(0.643,-0.498){67}{\rule{0.614pt}{0.120pt}}
\multiput(907.00,276.17)(43.725,-35.000){2}{\rule{0.307pt}{0.400pt}}
\multiput(952.00,240.92)(0.711,-0.497){59}{\rule{0.668pt}{0.120pt}}
\multiput(952.00,241.17)(42.614,-31.000){2}{\rule{0.334pt}{0.400pt}}
\multiput(996.00,209.92)(1.169,-0.495){35}{\rule{1.026pt}{0.119pt}}
\multiput(996.00,210.17)(41.870,-19.000){2}{\rule{0.513pt}{0.400pt}}
\multiput(1040.00,190.92)(1.523,-0.494){27}{\rule{1.300pt}{0.119pt}}
\multiput(1040.00,191.17)(42.302,-15.000){2}{\rule{0.650pt}{0.400pt}}
\multiput(1085.00,175.92)(2.052,-0.492){19}{\rule{1.700pt}{0.118pt}}
\multiput(1085.00,176.17)(40.472,-11.000){2}{\rule{0.850pt}{0.400pt}}
\multiput(1129.00,164.92)(2.267,-0.491){17}{\rule{1.860pt}{0.118pt}}
\multiput(1129.00,165.17)(40.139,-10.000){2}{\rule{0.930pt}{0.400pt}}
\multiput(1173.00,154.93)(3.391,-0.485){11}{\rule{2.671pt}{0.117pt}}
\multiput(1173.00,155.17)(39.455,-7.000){2}{\rule{1.336pt}{0.400pt}}
\multiput(1218.00,147.94)(6.330,-0.468){5}{\rule{4.500pt}{0.113pt}}
\multiput(1218.00,148.17)(34.660,-4.000){2}{\rule{2.250pt}{0.400pt}}
\multiput(1262.00,143.93)(4.829,-0.477){7}{\rule{3.620pt}{0.115pt}}
\multiput(1262.00,144.17)(36.487,-5.000){2}{\rule{1.810pt}{0.400pt}}
\put(1306,138.17){\rule{8.900pt}{0.400pt}}
\multiput(1306.00,139.17)(25.528,-2.000){2}{\rule{4.450pt}{0.400pt}}
\put(1350,136.17){\rule{9.100pt}{0.400pt}}
\multiput(1350.00,137.17)(26.112,-2.000){2}{\rule{4.550pt}{0.400pt}}
\put(1395,134.67){\rule{10.600pt}{0.400pt}}
\multiput(1395.00,135.17)(22.000,-1.000){2}{\rule{5.300pt}{0.400pt}}
\put(110,132){\makebox(0,0){$\bullet$}}
\put(154,133){\makebox(0,0){$\bullet$}}
\put(199,134){\makebox(0,0){$\bullet$}}
\put(243,137){\makebox(0,0){$\bullet$}}
\put(287,142){\makebox(0,0){$\bullet$}}
\put(332,154){\makebox(0,0){$\bullet$}}
\put(376,175){\makebox(0,0){$\bullet$}}
\put(420,215){\makebox(0,0){$\bullet$}}
\put(464,293){\makebox(0,0){$\bullet$}}
\put(509,422){\makebox(0,0){$\bullet$}}
\put(553,609){\makebox(0,0){$\bullet$}}
\put(597,707){\makebox(0,0){$\bullet$}}
\put(642,708){\makebox(0,0){$\bullet$}}
\put(686,640){\makebox(0,0){$\bullet$}}
\put(730,549){\makebox(0,0){$\bullet$}}
\put(775,465){\makebox(0,0){$\bullet$}}
\put(819,387){\makebox(0,0){$\bullet$}}
\put(863,327){\makebox(0,0){$\bullet$}}
\put(907,277){\makebox(0,0){$\bullet$}}
\put(952,242){\makebox(0,0){$\bullet$}}
\put(996,211){\makebox(0,0){$\bullet$}}
\put(1040,192){\makebox(0,0){$\bullet$}}
\put(1085,177){\makebox(0,0){$\bullet$}}
\put(1129,166){\makebox(0,0){$\bullet$}}
\put(1173,156){\makebox(0,0){$\bullet$}}
\put(1218,149){\makebox(0,0){$\bullet$}}
\put(1262,145){\makebox(0,0){$\bullet$}}
\put(1306,140){\makebox(0,0){$\bullet$}}
\put(1350,138){\makebox(0,0){$\bullet$}}
\put(1395,136){\makebox(0,0){$\bullet$}}
\put(1439,135){\makebox(0,0){$\bullet$}}
\put(1349,733){\makebox(0,0){$\bullet$}}
\put(1279,793){\makebox(0,0)[r]{$\mathcal{T}=Hyp$}} %%%%%%%%%%%%%%%%
\put(1299.0,793.0){\rule[-0.200pt]{24.090pt}{0.400pt}} %%%%%%%%%%%%%%%%
\put(110,132){\usebox{\plotpoint}}
\put(154,131.67){\rule{10.840pt}{0.400pt}}
\multiput(154.00,131.17)(22.500,1.000){2}{\rule{5.420pt}{0.400pt}}
\put(199,132.67){\rule{10.600pt}{0.400pt}}
\multiput(199.00,132.17)(22.000,1.000){2}{\rule{5.300pt}{0.400pt}}
\multiput(243.00,134.59)(4.829,0.477){7}{\rule{3.620pt}{0.115pt}}
\multiput(243.00,133.17)(36.487,5.000){2}{\rule{1.810pt}{0.400pt}}
\multiput(287.00,139.59)(2.592,0.489){15}{\rule{2.100pt}{0.118pt}}
\multiput(287.00,138.17)(40.641,9.000){2}{\rule{1.050pt}{0.400pt}}
\multiput(332.00,148.58)(1.109,0.496){37}{\rule{0.980pt}{0.119pt}}
\multiput(332.00,147.17)(41.966,20.000){2}{\rule{0.490pt}{0.400pt}}
\multiput(376.58,168.00)(0.498,0.533){85}{\rule{0.120pt}{0.527pt}}
\multiput(375.17,168.00)(44.000,45.906){2}{\rule{0.400pt}{0.264pt}}
\multiput(420.58,215.00)(0.498,1.060){85}{\rule{0.120pt}{0.945pt}}
\multiput(419.17,215.00)(44.000,91.038){2}{\rule{0.400pt}{0.473pt}}
\multiput(464.58,308.00)(0.498,1.888){87}{\rule{0.120pt}{1.602pt}}
\multiput(463.17,308.00)(45.000,165.675){2}{\rule{0.400pt}{0.801pt}}
\multiput(509.58,477.00)(0.498,2.699){85}{\rule{0.120pt}{2.245pt}}
\multiput(508.17,477.00)(44.000,231.339){2}{\rule{0.400pt}{1.123pt}}
\multiput(553.58,713.00)(0.498,1.038){85}{\rule{0.120pt}{0.927pt}}
\multiput(552.17,713.00)(44.000,89.075){2}{\rule{0.400pt}{0.464pt}}
\multiput(597.58,801.89)(0.498,-0.510){87}{\rule{0.120pt}{0.509pt}}
\multiput(596.17,802.94)(45.000,-44.944){2}{\rule{0.400pt}{0.254pt}}
\multiput(642.58,753.55)(0.498,-1.221){85}{\rule{0.120pt}{1.073pt}}
\multiput(641.17,755.77)(44.000,-104.773){2}{\rule{0.400pt}{0.536pt}}
\multiput(686.58,646.25)(0.498,-1.312){85}{\rule{0.120pt}{1.145pt}}
\multiput(685.17,648.62)(44.000,-112.623){2}{\rule{0.400pt}{0.573pt}}
\multiput(730.58,532.04)(0.498,-1.070){87}{\rule{0.120pt}{0.953pt}}
\multiput(729.17,534.02)(45.000,-94.021){2}{\rule{0.400pt}{0.477pt}}
\multiput(775.58,436.38)(0.498,-0.969){85}{\rule{0.120pt}{0.873pt}}
\multiput(774.17,438.19)(44.000,-83.189){2}{\rule{0.400pt}{0.436pt}}
\multiput(819.58,352.21)(0.498,-0.717){85}{\rule{0.120pt}{0.673pt}}
\multiput(818.17,353.60)(44.000,-61.604){2}{\rule{0.400pt}{0.336pt}}
\multiput(863.58,289.74)(0.498,-0.556){85}{\rule{0.120pt}{0.545pt}}
\multiput(862.17,290.87)(44.000,-47.868){2}{\rule{0.400pt}{0.273pt}}
\multiput(907.00,241.92)(0.683,-0.497){63}{\rule{0.645pt}{0.120pt}}
\multiput(907.00,242.17)(43.660,-33.000){2}{\rule{0.323pt}{0.400pt}}
\multiput(952.00,208.92)(0.849,-0.497){49}{\rule{0.777pt}{0.120pt}}
\multiput(952.00,209.17)(42.387,-26.000){2}{\rule{0.388pt}{0.400pt}}
\multiput(996.00,182.92)(1.489,-0.494){27}{\rule{1.273pt}{0.119pt}}
\multiput(996.00,183.17)(41.357,-15.000){2}{\rule{0.637pt}{0.400pt}}
\multiput(1040.00,167.92)(1.918,-0.492){21}{\rule{1.600pt}{0.119pt}}
\multiput(1040.00,168.17)(41.679,-12.000){2}{\rule{0.800pt}{0.400pt}}
\multiput(1085.00,155.93)(2.871,-0.488){13}{\rule{2.300pt}{0.117pt}}
\multiput(1085.00,156.17)(39.226,-8.000){2}{\rule{1.150pt}{0.400pt}}
\multiput(1129.00,147.93)(3.926,-0.482){9}{\rule{3.033pt}{0.116pt}}
\multiput(1129.00,148.17)(37.704,-6.000){2}{\rule{1.517pt}{0.400pt}}
\multiput(1173.00,141.95)(9.839,-0.447){3}{\rule{6.100pt}{0.108pt}}
\multiput(1173.00,142.17)(32.339,-3.000){2}{\rule{3.050pt}{0.400pt}}
\put(1218,138.17){\rule{8.900pt}{0.400pt}}
\multiput(1218.00,139.17)(25.528,-2.000){2}{\rule{4.450pt}{0.400pt}}
\multiput(1262.00,136.95)(9.616,-0.447){3}{\rule{5.967pt}{0.108pt}}
\multiput(1262.00,137.17)(31.616,-3.000){2}{\rule{2.983pt}{0.400pt}}
\put(1306,133.67){\rule{10.600pt}{0.400pt}}
\multiput(1306.00,134.17)(22.000,-1.000){2}{\rule{5.300pt}{0.400pt}}
\put(1350,132.67){\rule{10.840pt}{0.400pt}}
\multiput(1350.00,133.17)(22.500,-1.000){2}{\rule{5.420pt}{0.400pt}}
\put(110.0,132.0){\rule[-0.200pt]{10.600pt}{0.400pt}}
\put(110,132){\makebox(0,0){$\ast$}}
\put(154,132){\makebox(0,0){$\ast$}}
\put(199,133){\makebox(0,0){$\ast$}}
\put(243,134){\makebox(0,0){$\ast$}}
\put(287,139){\makebox(0,0){$\ast$}}
\put(332,148){\makebox(0,0){$\ast$}}
\put(376,168){\makebox(0,0){$\ast$}}
\put(420,215){\makebox(0,0){$\ast$}}
\put(464,308){\makebox(0,0){$\ast$}}
\put(509,477){\makebox(0,0){$\ast$}}
\put(553,713){\makebox(0,0){$\ast$}}
\put(597,804){\makebox(0,0){$\ast$}}
\put(642,758){\makebox(0,0){$\ast$}}
\put(686,651){\makebox(0,0){$\ast$}}
\put(730,536){\makebox(0,0){$\ast$}}
\put(775,440){\makebox(0,0){$\ast$}}
\put(819,355){\makebox(0,0){$\ast$}}
\put(863,292){\makebox(0,0){$\ast$}}
\put(907,243){\makebox(0,0){$\ast$}}
\put(952,210){\makebox(0,0){$\ast$}}
\put(996,184){\makebox(0,0){$\ast$}}
\put(1040,169){\makebox(0,0){$\ast$}}
\put(1085,157){\makebox(0,0){$\ast$}}
\put(1129,149){\makebox(0,0){$\ast$}}
\put(1173,143){\makebox(0,0){$\ast$}}
\put(1218,140){\makebox(0,0){$\ast$}}
\put(1262,138){\makebox(0,0){$\ast$}}
\put(1306,135){\makebox(0,0){$\ast$}}
\put(1350,134){\makebox(0,0){$\ast$}}
\put(1395,133){\makebox(0,0){$\ast$}}
\put(1439,133){\makebox(0,0){$\ast$}}
\put(1349,793){\makebox(0,0){$\ast$}} %%%%%%%%%%%%%%%%
\put(1395.0,133.0){\rule[-0.200pt]{10.600pt}{0.400pt}}
\put(110.0,131.0){\rule[-0.200pt]{0.400pt}{175.375pt}}
\put(110.0,131.0){\rule[-0.200pt]{320.156pt}{0.400pt}}
\put(1439.0,131.0){\rule[-0.200pt]{0.400pt}{175.375pt}}
\put(110.0,859.0){\rule[-0.200pt]{320.156pt}{0.400pt}}
\end{picture}

%% file: crossings.tex
% GNUPLOT: LaTeX picture
\setlength{\unitlength}{0.240900pt}
\ifx\plotpoint\undefined\newsavebox{\plotpoint}\fi
\sbox{\plotpoint}{\rule[-0.200pt]{0.400pt}{0.400pt}}%
\begin{picture}(1500,900)(0,0)
\sbox{\plotpoint}{\rule[-0.200pt]{0.400pt}{0.400pt}}%
\put(171.0,131.0){\rule[-0.200pt]{4.818pt}{0.400pt}}
\put(151,131){\makebox(0,0)[r]{-1}}
\put(1419.0,131.0){\rule[-0.200pt]{4.818pt}{0.400pt}}
\put(171.0,204.0){\rule[-0.200pt]{4.818pt}{0.400pt}}
\put(151,204){\makebox(0,0)[r]{-0.5}}
\put(1419.0,204.0){\rule[-0.200pt]{4.818pt}{0.400pt}}
\put(171.0,277.0){\rule[-0.200pt]{4.818pt}{0.400pt}}
\put(151,277){\makebox(0,0)[r]{ 0}}
\put(1419.0,277.0){\rule[-0.200pt]{4.818pt}{0.400pt}}
\put(171.0,349.0){\rule[-0.200pt]{4.818pt}{0.400pt}}
\put(151,349){\makebox(0,0)[r]{ 0.5}}
\put(1419.0,349.0){\rule[-0.200pt]{4.818pt}{0.400pt}}
\put(171.0,422.0){\rule[-0.200pt]{4.818pt}{0.400pt}}
\put(151,422){\makebox(0,0)[r]{ 1}}
\put(1419.0,422.0){\rule[-0.200pt]{4.818pt}{0.400pt}}
\put(171.0,495.0){\rule[-0.200pt]{4.818pt}{0.400pt}}
\put(151,495){\makebox(0,0)[r]{ 1.5}}
\put(1419.0,495.0){\rule[-0.200pt]{4.818pt}{0.400pt}}
\put(171.0,568.0){\rule[-0.200pt]{4.818pt}{0.400pt}}
\put(151,568){\makebox(0,0)[r]{ 2}}
\put(1419.0,568.0){\rule[-0.200pt]{4.818pt}{0.400pt}}
\put(171.0,641.0){\rule[-0.200pt]{4.818pt}{0.400pt}}
\put(151,641){\makebox(0,0)[r]{ 2.5}}
\put(1419.0,641.0){\rule[-0.200pt]{4.818pt}{0.400pt}}
\put(171.0,713.0){\rule[-0.200pt]{4.818pt}{0.400pt}}
\put(151,713){\makebox(0,0)[r]{ 3}}
\put(1419.0,713.0){\rule[-0.200pt]{4.818pt}{0.400pt}}
\put(171.0,786.0){\rule[-0.200pt]{4.818pt}{0.400pt}}
\put(151,786){\makebox(0,0)[r]{ 3.5}}
\put(1419.0,786.0){\rule[-0.200pt]{4.818pt}{0.400pt}}
\put(171.0,859.0){\rule[-0.200pt]{4.818pt}{0.400pt}}
\put(151,859){\makebox(0,0)[r]{ 4}}
\put(1419.0,859.0){\rule[-0.200pt]{4.818pt}{0.400pt}}
\put(171.0,131.0){\rule[-0.200pt]{0.400pt}{4.818pt}}
\put(171,90){\makebox(0,0){ 0}}
\put(171.0,839.0){\rule[-0.200pt]{0.400pt}{4.818pt}}
\put(312.0,131.0){\rule[-0.200pt]{0.400pt}{4.818pt}}
\put(312,90){\makebox(0,0){ 1}}
\put(312.0,839.0){\rule[-0.200pt]{0.400pt}{4.818pt}}
\put(453.0,131.0){\rule[-0.200pt]{0.400pt}{4.818pt}}
\put(453,90){\makebox(0,0){ 2}}
\put(453.0,839.0){\rule[-0.200pt]{0.400pt}{4.818pt}}
\put(594.0,131.0){\rule[-0.200pt]{0.400pt}{4.818pt}}
\put(594,90){\makebox(0,0){ 3}}
\put(594.0,839.0){\rule[-0.200pt]{0.400pt}{4.818pt}}
\put(735.0,131.0){\rule[-0.200pt]{0.400pt}{4.818pt}}
\put(735,90){\makebox(0,0){ 4}}
\put(735.0,839.0){\rule[-0.200pt]{0.400pt}{4.818pt}}
\put(875.0,131.0){\rule[-0.200pt]{0.400pt}{4.818pt}}
\put(875,90){\makebox(0,0){ 5}}
\put(875.0,839.0){\rule[-0.200pt]{0.400pt}{4.818pt}}
\put(1016.0,131.0){\rule[-0.200pt]{0.400pt}{4.818pt}}
\put(1016,90){\makebox(0,0){ 6}}
\put(1016.0,839.0){\rule[-0.200pt]{0.400pt}{4.818pt}}
\put(1157.0,131.0){\rule[-0.200pt]{0.400pt}{4.818pt}}
\put(1157,90){\makebox(0,0){ 7}}
\put(1157.0,839.0){\rule[-0.200pt]{0.400pt}{4.818pt}}
\put(1298.0,131.0){\rule[-0.200pt]{0.400pt}{4.818pt}}
\put(1298,90){\makebox(0,0){ 8}}
\put(1298.0,839.0){\rule[-0.200pt]{0.400pt}{4.818pt}}
\put(1439.0,131.0){\rule[-0.200pt]{0.400pt}{4.818pt}}
\put(1439,90){\makebox(0,0){ 9}}
\put(1439.0,839.0){\rule[-0.200pt]{0.400pt}{4.818pt}}
\put(171.0,131.0){\rule[-0.200pt]{0.400pt}{175.375pt}}
\put(171.0,131.0){\rule[-0.200pt]{305.461pt}{0.400pt}}
\put(1439.0,131.0){\rule[-0.200pt]{0.400pt}{175.375pt}}
\put(171.0,859.0){\rule[-0.200pt]{305.461pt}{0.400pt}}
\put(30,495){\makebox(0,0){\%}}
\put(805,29){\makebox(0,0){Number of crossed alignments}}
\put(171,816){\usebox{\plotpoint}}
\multiput(171.58,808.14)(0.499,-2.245){279}{\rule{0.120pt}{1.893pt}}
\multiput(170.17,812.07)(141.000,-628.071){2}{\rule{0.400pt}{0.946pt}}
\multiput(312.00,184.59)(10.714,0.485){11}{\rule{8.157pt}{0.117pt}}
\multiput(312.00,183.17)(124.069,7.000){2}{\rule{4.079pt}{0.400pt}}
\multiput(594.00,191.58)(2.294,0.497){59}{\rule{1.919pt}{0.120pt}}
\multiput(594.00,190.17)(137.016,31.000){2}{\rule{0.960pt}{0.400pt}}
\put(453.0,191.0){\rule[-0.200pt]{33.967pt}{0.400pt}}
\multiput(875.00,222.58)(2.973,0.496){45}{\rule{2.450pt}{0.120pt}}
\multiput(875.00,221.17)(135.915,24.000){2}{\rule{1.225pt}{0.400pt}}
\put(1016,245.67){\rule{33.967pt}{0.400pt}}
\multiput(1016.00,245.17)(70.500,1.000){2}{\rule{16.983pt}{0.400pt}}
\multiput(1157.00,247.59)(9.277,0.488){13}{\rule{7.150pt}{0.117pt}}
\multiput(1157.00,246.17)(126.160,8.000){2}{\rule{3.575pt}{0.400pt}}
\multiput(1298.00,255.59)(10.714,0.485){11}{\rule{8.157pt}{0.117pt}}
\multiput(1298.00,254.17)(124.069,7.000){2}{\rule{4.079pt}{0.400pt}}
\put(171,816){\makebox(0,0){$\bullet$}}
\put(312,184){\makebox(0,0){$\bullet$}}
\put(453,191){\makebox(0,0){$\bullet$}}
\put(594,191){\makebox(0,0){$\bullet$}}
\put(735,222){\makebox(0,0){$\bullet$}}
\put(875,222){\makebox(0,0){$\bullet$}}
\put(1016,246){\makebox(0,0){$\bullet$}}
\put(1157,247){\makebox(0,0){$\bullet$}}
\put(1298,255){\makebox(0,0){$\bullet$}}
\put(1439,262){\makebox(0,0){$\bullet$}}
\put(735.0,222.0){\rule[-0.200pt]{33.726pt}{0.400pt}}
\put(171,277){\usebox{\plotpoint}}
\put(171.00,277.00){\usebox{\plotpoint}}
\put(191.76,277.00){\usebox{\plotpoint}}
\put(212.51,277.00){\usebox{\plotpoint}}
\put(233.27,277.00){\usebox{\plotpoint}}
\put(254.02,277.00){\usebox{\plotpoint}}
\put(274.78,277.00){\usebox{\plotpoint}}
\put(295.53,277.00){\usebox{\plotpoint}}
\put(316.29,277.00){\usebox{\plotpoint}}
\put(337.04,277.00){\usebox{\plotpoint}}
\put(357.80,277.00){\usebox{\plotpoint}}
\put(378.55,277.00){\usebox{\plotpoint}}
\put(399.31,277.00){\usebox{\plotpoint}}
\put(420.07,277.00){\usebox{\plotpoint}}
\put(440.82,277.00){\usebox{\plotpoint}}
\put(461.58,277.00){\usebox{\plotpoint}}
\put(482.33,277.00){\usebox{\plotpoint}}
\put(503.09,277.00){\usebox{\plotpoint}}
\put(523.84,277.00){\usebox{\plotpoint}}
\put(544.60,277.00){\usebox{\plotpoint}}
\put(565.35,277.00){\usebox{\plotpoint}}
\put(586.11,277.00){\usebox{\plotpoint}}
\put(606.87,277.00){\usebox{\plotpoint}}
\put(627.62,277.00){\usebox{\plotpoint}}
\put(648.38,277.00){\usebox{\plotpoint}}
\put(669.13,277.00){\usebox{\plotpoint}}
\put(689.89,277.00){\usebox{\plotpoint}}
\put(710.64,277.00){\usebox{\plotpoint}}
\put(731.40,277.00){\usebox{\plotpoint}}
\put(752.15,277.00){\usebox{\plotpoint}}
\put(772.91,277.00){\usebox{\plotpoint}}
\put(793.66,277.00){\usebox{\plotpoint}}
\put(814.42,277.00){\usebox{\plotpoint}}
\put(835.18,277.00){\usebox{\plotpoint}}
\put(855.93,277.00){\usebox{\plotpoint}}
\put(876.69,277.00){\usebox{\plotpoint}}
\put(897.44,277.00){\usebox{\plotpoint}}
\put(918.20,277.00){\usebox{\plotpoint}}
\put(938.95,277.00){\usebox{\plotpoint}}
\put(959.71,277.00){\usebox{\plotpoint}}
\put(980.46,277.00){\usebox{\plotpoint}}
\put(1001.22,277.00){\usebox{\plotpoint}}
\put(1021.98,277.00){\usebox{\plotpoint}}
\put(1042.73,277.00){\usebox{\plotpoint}}
\put(1063.49,277.00){\usebox{\plotpoint}}
\put(1084.24,277.00){\usebox{\plotpoint}}
\put(1105.00,277.00){\usebox{\plotpoint}}
\put(1125.75,277.00){\usebox{\plotpoint}}
\put(1146.51,277.00){\usebox{\plotpoint}}
\put(1167.26,277.00){\usebox{\plotpoint}}
\put(1188.02,277.00){\usebox{\plotpoint}}
\put(1208.77,277.00){\usebox{\plotpoint}}
\put(1229.53,277.00){\usebox{\plotpoint}}
\put(1250.29,277.00){\usebox{\plotpoint}}
\put(1271.04,277.00){\usebox{\plotpoint}}
\put(1291.80,277.00){\usebox{\plotpoint}}
\put(1312.55,277.00){\usebox{\plotpoint}}
\put(1333.31,277.00){\usebox{\plotpoint}}
\put(1354.06,277.00){\usebox{\plotpoint}}
\put(1374.82,277.00){\usebox{\plotpoint}}
\put(1395.57,277.00){\usebox{\plotpoint}}
\put(1416.33,277.00){\usebox{\plotpoint}}
\put(1437.09,277.00){\usebox{\plotpoint}}
\put(1439,277){\usebox{\plotpoint}}
\put(171.0,131.0){\rule[-0.200pt]{0.400pt}{175.375pt}}
\put(171.0,131.0){\rule[-0.200pt]{305.461pt}{0.400pt}}
\put(1439.0,131.0){\rule[-0.200pt]{0.400pt}{175.375pt}}
\put(171.0,859.0){\rule[-0.200pt]{305.461pt}{0.400pt}}
\end{picture}